\documentclass[conference]{IEEEtran}
\IEEEoverridecommandlockouts
\usepackage{cite}
\usepackage{amsmath,amssymb,amsfonts}
\usepackage{algorithmic}
\usepackage{graphicx}
\usepackage{textcomp}
\usepackage{xcolor}
\usepackage{hyperref}
\usepackage{multirow}

\def\BibTeX{{\rm B\kern-.05em{\sc i\kern-.025em b}\kern-.08em
    T\kern-.1667em\lower.7ex\hbox{E}\kern-.125emX}}
\begin{document}

\title{QuickBrowser: A Unified Model to Detect and Read Simple Object in Real-time}

\author{\IEEEauthorblockN{Thao Do}
\IEEEauthorblockA{KAIST \\
Daejeon, South Korea \\
thaodo@kaist.ac.kr}
\and
\IEEEauthorblockN{Daeyoung Kim}
\IEEEauthorblockA{KAIST \\
Daejeon, South Korea \\
kimd@kaist.ac.kr}
}

\maketitle

\begin{abstract}
There are many real-life use cases such as barcode scanning or billboard reading where people need to detect objects and read the object contents. Commonly existing methods are first trying to localize object regions, then determine layout and lastly classify content units. However, for simple fixed structured objects like license plates, this approach becomes surplus and heavy to run. This work aims to solve this detect-and-read problem in a lightweight way by integrating multi-digit recognition into a one-stage object detection model. Our unified method not only eliminates the duplication in feature extraction (one for localizing, one again for classifying) but also provides useful contextual information around object regions for classification. Additionally, our choice of backbones and modifications in architecture, loss function, data augmentation and training make the method robust, efficient and speedy. Secondly, we made a public benchmark dataset of diverse real-life 1D barcodes for a reliable evaluation, which we collected, annotated and checked carefully. Eventually, experimental results prove the method's efficiency on the barcode problem by outperforming industrial tools in both detecting and decoding rates with a real-time fps at a VGA-similar resolution. It also did a great job expectedly on the license-plate recognition task (on the AOLP dataset) by outperforming the current state-of-the-art method significantly in terms of recognition rate and inference time.
\end{abstract}


\section{Introduction}
Deep convolutional neural networks have recently shown remarkable performance in vision-related tasks such as image classification, object detection, or segmentation. From those successes, we witnessed many applications utilizing those fundamental advances to tackle practical, useful jobs, although their scope and accuracy are still limited. One of the typical use cases is that we need to look for objects and read the object contents. For example, we're walking around a street to browse billboards, panels and also read to contents of those objects for next steps like making a phone call by retrieved numbers or surfing their website by URL. As a norm, most machine approaches tackling this kind of application are step-by-step: using an object detector to detect the regions of interest (ROI) and then read the object clearly by another model. This way proved its robustness in many problems, especially when the items are very varied, non-standardized, and have highly complex structured content such as clothing labels, shop signs, etc. However, applying this stereotype to simple, standardized objects is not flexible and lengthy. Two typical examples of simple items are one-dimensional barcodes and license plates. 

Barcode scanning in supermarkets or warehouse is daily, time-consuming and unhandy work for both customers and staff these days. Such devices assisting workers to pick tagged packages correctly would be a beneficial applicable scenario. In this work, we mainly talk about camera-based barcode scanning (to differentiate from laser-based solutions). Many approaches proposed in 20 past years are getting better and better in performance and speed; however, they are still limited and have not yet reach real-time speed. Briefly, existing works could be categorized into 2 types: localizing focus and decoding focus, which are also 2 sub-tasks of the problem under the view of conventional approaches. Localizing-focus works tried to get a better detection rate under various real conditions like blurry, low-light, and other kinds of noise. On the other hand, decoding-focus works concentrate on translating the localized (or preprocessed) regions from the localizing task to encoded sequences correctly. 

Both sub-tasks were quite challenging and required handcrafted feature extraction as in \cite{joseph1994bar, muniz1999robust, wachenfeld2008robust} until the surging of the deep convolutional neural network (DNN). DNN-era approaches as \cite{fridborn2017reading} use a CNN-based multi-label classifier to decode all digits of a barcode sequence at once proved robustness even in challenging conditional inputs whereas \cite{hansen2017real} attacked mainly on detection by adopted a DNN-based object detector to reach high detection rate surpassing all previous works \cite{lin2011real,katona2012novel,soros2013blur,creusot2015real,creusot2016low}. Although recent works almost conquer each sub-task, the end-to-end task naively joining the two sub-tasks becomes slow and thus non-real time because of this pipeline structure and the duplication in extracting features.

Similarly, license plates in most countries are often standardized designed with a limited maximum number of characters and symbols, while Automatic License Plate Recognition (ALPR) has many useful application domains in traffic controlling and parking management. Conventional way divides end-to-end task to 3 step-by-step tasks: object detection, segmentation, and optical character recognition as \cite{safaei2016real,kumar2016efficient}. In 2019, \cite{chen2019automatic} proposed a faster methodology by using an ensemble of multiple tiny letter detectors and combining reasonably those sub-results to final results. \cite{chen2019automatic} boosted its speed by skipping segmentation but it adds a logical aggregation of those tiny model outputs. Though tiny detectors work well independently, they bring many noise characters (out-of-the-plate characters). Final aggregation takes a considerable time to validate characters, dismiss noises, arrange letter orders, and group sub-results to the final boxes. So mainly the method was time-consuming and far from real-time performance. 

Therefore, in this study, we proposed a unified model to deal with the end-to-end detect-and-read-content problem for simple structured objects with the following contributions:
\begin{itemize}
\item Our method -- QuickBrowser successfully integrates multi-digit classification into DNN one-stage object detection along with modified architecture, loss function, data augmentation, and training strategy, making it fast, computational saving, extensible and robust even under various harsh conditions and object smallness.
\item Experimental results on our reliable barcode dataset and the AOLP license plate dataset demonstrated the method's efficiency by outperforming industrial software, existing works in terms of detection rate, recognition rate, and inference time while our fastest models achieve real-time and near-real-time speed under practical resolutions.
\item Lastly, we extended and upgraded the \cite{do2020smart} dataset to a more standard barcode dataset by collecting more samples, annotating, and double-checking. The dataset has 3 subsets: a main EAN-13 single-barcode set having a clear train/validation/test split, an EAN-13 multi-barcode set, and a small EAN-8 set to confirm the extensibility. The dataset is public to challenge further research.
\end{itemize}

\section{Related Work}

Regarding barcodes - a typical example of simple objects aforementioned, there are two development flows: localization and decoding. In the first flow, \cite{lin2011real} introduced the first multiple and rotation-tolerated barcode recognition method in 2011. The work uses some traditional filters to segment out barcode regions, enhanced the stripes, rotated the regions to horizontal before putting them to a regular decoder. Though the method worked well on plain papers (lottery tickets), it was still slow and no other material results. \cite{katona2012novel} proposed a method using morphological operations to segment out barcode regions under various conditions with good performance in 2012. \cite{soros2013blur} in 2013 focused on dealing with blur using structure matrix and saturation from HSV color system to detect better but with the cost of speed. Recently, \cite{creusot2016low} offered a faster method for also blurred input based on Line Segment Detector following their previous work \cite{creusot2015real} using Maximal Stable Extremal Region shown sensitive to blur. Lastly, \cite{hansen2017real} in 2017 first applied a DNN object detection, shown achieving the best detection rate on small datasets at that time expectedly. \cite{hansen2017real} also tried to predict horizontal-rotating angles by another model and input cropped rotated barcode patches into a traditional decoder. A common point most previous works sharing is that they still need a decoder to complete their final goal.

On the decoding flow, several works had been proposed sparsely since the 1990s. Early works \cite{joseph1994bar,muniz1999robust} achieved their goal by techniques as Hough transformation, wavelet-based peak location on their naive inputs. In 2008, \cite{wachenfeld2008robust} proposed a scanline-based approach which just worked well on a slightly rotated (\textpm 15 degrees) barcode; stronger rotated or distorted would be problematic. Similar to \cite{katona2012novel}, \cite{zamberletti2010decoding} also attacked out-of-focus input by using a multi-layer perceptron model to find an adaptive threshold to enhance blurry patches to clearer before decoding by a regular decoder. Recently, \cite{yang2016automatic} proposed a method heavily based on scanline-base and hand-crafty featuring and analysis for each challenging condition. Despite getting high detection and recognition rate, it is a less scalable solution to extend other 1D types. 
More recently, \cite{fridborn2017reading} in 2017 leveraged CNN to directly extract features and predict simultaneously all digits of barcode sequence given a localized region. The idea of this method was actually coined in \cite{goodfellow2013multi} so we call it \emph{multi-digit} classification. Compared with its previous methods, the DNN-based approach is straightforward and data-driven than case-by-case analysis. One noticeable case when there is no single line crossing all the code bars is problematic for scanline-based approaches, but it could be learned and overcome by a DNN classifier. \cite{do2020smart} followed that CNN base but improved it by exploiting the self-validation feature of the barcode and test-time augmentation. Eventually, decoders need locators as their inputs are barcode-focus cropped patches. This loose-coupling has advantages in some cases but with the cost of speed. Thus, our method, which its application scope focuses more on speed, is mainly tight-coupled by its integrated design.

Another example of simple objects we try to apply our method on is vehicle license plate detection and recognition. The topic attracted numerous research conducted in \cite{anagnostopoulos2008license,sheng2009real,hsu2012application,chen2019automatic}. The plate detection predicts bounding boxes for the plate regions given the wild image. Conventional approaches that use morphological categorizations could be grouped in some ways: edge-based, color-based, texture-based, and character-based. Edge-based \cite{anagnostopoulos2008license} methods highlight the edges of black characters against the plate's white background and the plate frame with the vehicle body. Similarly, \cite{sheng2009real} briefly reduces noise after edge-based extraction from the license plate. Despite fast processing, it is sensitive to unwanted edges. Color-based approaches basically emphasize the difference between vehicle body color and plate color, such as \cite{deb2008hsi} proposed an HSI color model to detect candidate regions, which is later filtered out by histogram. The color-based could detect inclined and deformed license plates; however, it failed if the vehicle body and plate colors are similar or under the illumination changes in outdoor locations.
Unlike the edge or color approaches, \cite{hsu2012application} proposed a texture-based method that extracts dense sets and rectangular shapes similar to a  plate shape and then verifies them by Gaussian mixture model with Expectation-Maximization algorithm. Features extracted from this way are more discriminative but with the cost of computation, time. Besides, it still handcraft features that mean takes time when new types and shapes are added. Finally, character-based approaches powered by CNN detect the objects using the string of characters extracted directly from the image such as \cite{chen2019automatic} we mentioned in the Introduction. The last type of approach is potential because of automatically sophisticated feature learning and the reduction step of reading plate sequences. Although \cite{chen2019automatic} took the state-of-the-art place in AOLP dataset in 2019, there are some limitation: maintaining a bunch of tiny models is inconvenient in deployments, improving YOLO to deal with small objects by sliding windows is quite time-consuming; and final aggregation takes a nontrivial period as summarised in the penultimate part of Introduction.

From a modern view, simple object localization (i.e., detection) is just a simplified case of generic object detection. DNN-based object detectors were born gradually to conquer higher accuracy, more extensive scale. There are mainly two types of DNN-adopted models: one-stage and two-stage. Typical two-stage models are the R-CNN family \cite{girshick2014rich} where detection happens in two stages: first, the model proposes a set of regions of interests by selective search or regional proposal network (bounding box candidates can be infinite), then the classifier part classifies the region candidates.
On the other hand, one-stage models like SSD \cite{liu2016ssd}, or YOLO \cite{redmon2016you} skips the proposal stage and detect directly on a dense sampling of possible locations. This is quicker and simpler but potentially with the cost of accuracy \cite{liu2020deep}. Methods proposed or improved in recent years getting more powerful and complicated could be overkill for simple cases. YOLO's approach is straightforward but yet effective, rapid and widespread adopted; hence, we adopt this head in our architecture in this work. Aforesaid, a common limitation of most of the existing approaches is that breaking the end-to-end task of detect-and-read-content into 2-stage tasks which slows down the process while disposing valuable extracted feature from the localization.

\section{Methodology}

We focus on solving this end-to-end problem by unifying multi-digit classification (as \cite{goodfellow2013multi}) and DNN one-stage object detection. To the best of our knowledge, this is the first work trying to incorporate one-stage object detection with multi-digit CNN to deal with the problem for simple objects.

\subsection{Unified Recognition}
Without losing the generalization, we opt for YOLO \cite{redmon2016you} head model due to its simplicity and the task suitability. Compared to YOLO version 2,3 \cite{redmon2017yolo9000, redmon2018yolov3} or SSD \cite{liu2016ssd}, the use of anchor boxes (i.e., box priors) is unnecessary since barcode is omnidirectional (i.e., readers must detect and read no matter rotation) and stretchable which cause height-width ratios are not clustered like what happens to datasets like COCO \cite{lin2014microsoft}. Furthermore, recent models trying to solve absolute overlapping objects (e.g., predict multi objects per cell) with the higher cost of computation while overlapped barcodes are non-decodable and things like plate-on-plate cases are non-existed or deliberated law-violated.

Similar to YOLO \cite{redmon2016you}, our system models the task as a regression problem. The input is divided into an $S \times S$ grid, and one cell is responsible for detecting an object if that object center lay inside the cell. Each cell predicts $B$ bounding boxes and confidence scores for those boxes. Specifically, each box of $B$ bounding boxes predicts 5 figures: $x, y, w, h$ (scaled, relative to the cell by offsetting) and $confidence$. However, in classification, instead of solely predicting (only one or) a few object classes, we predict in the same way as multi-digit CNN did in\cite{goodfellow2013multi}. In other words, we \emph{read} an object content simply by viewing it as a single structured sequence and simultaneously predict $L$ digits ($L$ is the maximum/fixed length of the sequence $D$) and the real length $l$ of the sequence (if length is not fixed). Each digit $i$ is predicted through $N$ conditional class probabilities $Pr(j\_D[i]|Object)$ (e.g. 1D barcode EAN-13 has $L=13$ digits, each digit could be one of $N=10(11)$ values from $j=0$ to $9$ and $not$ $available$ if necessary). Furthermore, as the condition of no absolute overlap, we decided to make the grid denser and design each cell only represents only one sequence regardless of the number of boxes $B$.

Finally, at the test time, we take the product of box confidence and arithmetic mean of all variables' max probabilities, $confidence \times \frac{1}{L} \sum_{i}^{L} max(Pr(j\_D[i]|Object))$ and this final confidence reflexes how well that predicted box locate the object and the confidence of the model in predicting the sequence correctly. Those boxes with final confidences are then inputted into non-maximum suppression (NMS) with a threshold to reduce overlapping boxes that predict same objects. The after-NMS sequences satisfying validity (e.g. barcode satisfied checksum, predicted plate length agreed with $l$ -- length variable) are reported, while invalid ones but with high confidence are still listed as unreadable objects needing adjustments such as zoom in or hold still, etc.

\subsection{Network Architecture}
Overall, implemented model has 3 parts: backbone, neck and head as shown in Figure \ref{fig:end_architect} below. We take extraction parts of ResNet \cite{he2016deep} and MobileNetV2 \cite{sandler2018mobilenetv2} as backbones instead of Darknet19 \cite{redmon2017yolo9000} because papers report that ResNet backbones (with skip connections) gave the better results in term of accuracy \cite{redmon2018yolov3} while MobileNetV2 pays more attention to speed while still maintaining good accuracy \cite{sandler2018mobilenetv2}.

On top of that, we put a stack of Dilated BottleNeck blocks (proposed in \cite{yu2017dilated}) in a similar way to DetNet \cite{li2018detnet} which demonstrated maintaining the spatial resolution and enlarging the receptive field, which is good for the case of smaller objects are often dominated by bigger objects. This is valuable since object content texture (e.g., barcode bars) are individually small in wild images. Note that this stack slows down the model a bit by increasing its complexity. Eventually, a batch normalization followed by an adaptive max pooling and a 2D convolution are taped on the neck to normalize and ensure the shape of the output follows what we want no matter the input size.

\begin{figure*}[ht]
\centering
  \includegraphics[width=\textwidth,keepaspectratio]{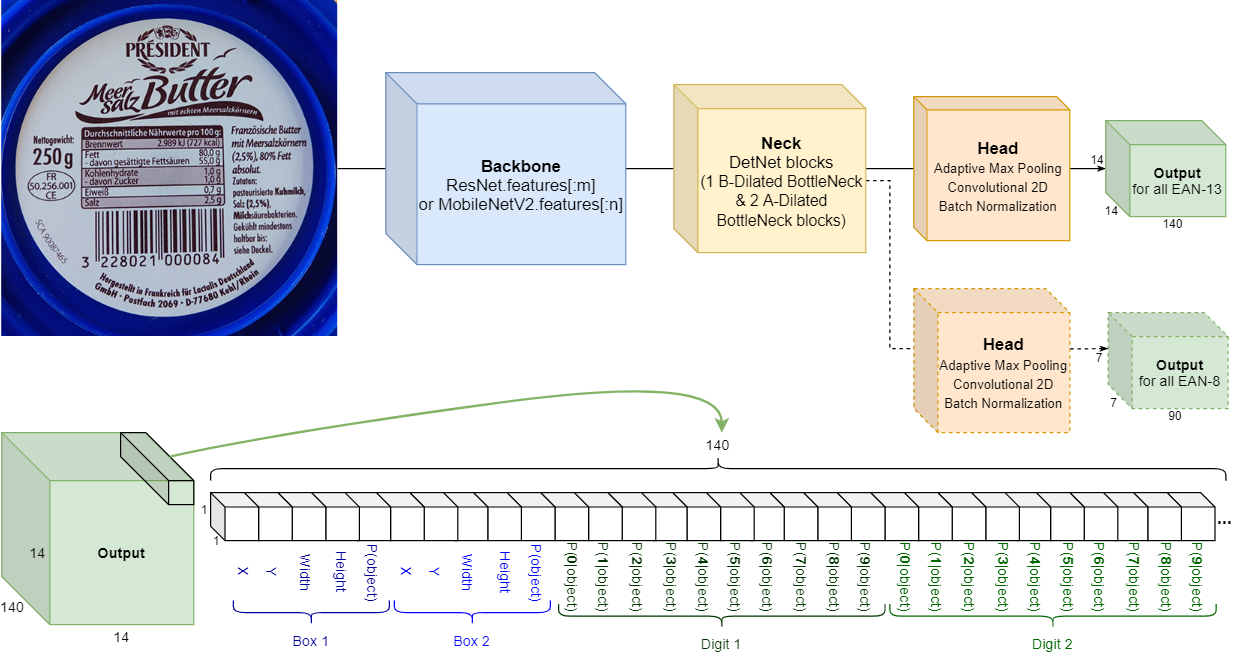}
  \caption{A general architecture for end-to-end recognition}
  \label{fig:end_architect}
\end{figure*}

This type of model not only accelerates the end-to-end process by making use of features extracted from DNN but also gives a better prediction as it provides the classification part of the model supportive contextual information around the regions of interest. Intuitively, there might be not enough clue to identify too small objects but cropped without context around them or the object image is under non-uniform conditions (non-uniform motion blur or non-uniform cover).

\subsection{Training}
\begin{equation}
\label{loss_func}
\begin{split}
&\lambda_{coord} \sum_{i=0}^{S^2}\sum_{j=0}^B \textbf{1}_{ij}^{obj}[(x_i-\hat{x}_i)^2 + (y_i-\hat{y}_i)^2 ] \\&+ \lambda_{coord} \sum_{i=0}^{S^2}\sum_{j=0}^B \textbf{1}_{ij}^{obj}[(\sqrt{w_i}-\sqrt{\hat{w}_i})^2 +(\sqrt{h_i}-\sqrt{\hat{h}_i})^2 ]\\
&+ \sum_{i=0}^{S^2}\sum_{j=0}^B \textbf{1}_{ij}^{obj}(C_i - \hat{C}_i)^2 + \lambda_{noobj}\sum_{i=0}^{S^2}\sum_{j=0}^B \textbf{1}_{ij}^{noobj}(C_i - \hat{C}_i)^2 \\
&+ \mathbf{\lambda_{class}} \sum_{i=0}^{S^2} \textbf{1}_{i}^{obj}\sum_{c \in digits}(p_i(c) - \hat{p}_i(c))^2
\end{split}
\end{equation}

Using the YOLO \cite{redmon2016you} loss function defined as the expression \ref{loss_func} below. The first 2 lines represent the loss of bounding box regression part (the differences between the ground-truth and the predicted of box's center coordinates on the first line, box's size on the second line) while the third line describes the loss of object confidence part. However, in the last part which is classification loss by mean square error, we add \scalebox{1.2}{$\mathbf{\lambda_{class}}$} to weight the importance of predicting content sequence correctly. Furthermore, experiments showed that this weight useful in boosting model recognition rate; however, we have to carefully set and adjust this hyper-parameter value along with the learning rate decay schedule, mixing real/synthesized ratio during training to achieve the best results.

\subsection{Data Augmentation \& Synthesizing Data}

Despite the similar structure, data augmentation applied to each object should be considered based on the object characteristic. For example, barcodes are omnidirectional and sensitive to horizontal loss (cause non-decodable), so besides common augmentations (like make blur, adjust brightness/hue/saturation, flip horizontally, add noise), we use various barcode-centric augmentations: random cropping/shifting without cutting off bounding boxes, random rotating within a circle (\($360\textdegree$\)), stretching, synthesizing elastic effects \cite{simard2003best} (to make it resemble a wrinkled barcode on plastic bags), shearing randomly both horizontally and vertically. On the other hand, license plates are different because Arabic numerals are sensitive to strong rotation or flipping (e.g., 6 becomes 9 after \($180\textdegree$\) rotation, 3 becomes a weird symbol).

Due to the limited number of real samples, synthesizing simple objects which is not complicated is extremely beneficial for model training to be converged. In the barcode case, numerous random barcode sequences complied with naming convention are generated, encoded to barcode form. They are then transformed perspective (3D rotated) or augmented randomly before taped on random images. Similarly, large synthesized license plates are created by randomizing sequences on random vehicle plate templates by a font similar to the real plate's font. Those plates are undergone random perspective transform and noise adding operations to make them more realistic. The detailed processes and their effects are described in the Experiments part.

\section{Experiments}
\subsection{Datasets}

We prepared a trustable real dataset for the barcode recognition that has 3 standard subsets: training set, validation set, and test set as usual. The dataset was extended from the \cite{do2020smart} work, and our new collecting campaign (also carefully double-check and annotate bounding box) from real image captures and 2 crawling online on product tagging websites: \href{https://world.openfoodfacts.org/}{world.openfoodfacts.org} \& \href{https://www.foodrepo.org/}{foodrepo.org} \cite{lazzari2018foodrepo}. Those images are EAN-13 single-barcode and various in size and format. Totally we have $3200$ samples comprised of $2000$ samples for testing, $900$ samples for training, and $300$ ones for validation. We limit our primary test dataset at single-barcode images because most crawling samples are single-barcode and the LabelImage tool we used to annotate is not ready for writing multiple decoded sequences. Eventually, we also prepared 2 small sets of 20 multi-barcode images ($77$ barcodes total) and 90 EAN-8 samples ($0$ for training, $90$ for evaluation) for supporting demonstrations.

Barcodes are synthesized with a variety of height ($3-20$), width ($7-11$), quiet zone ($1-6$), valid background and foreground colors, gap widths, and no numerals printed to force models to learn strip pattern, not numerals. We paste $0 - 3$ barcodes per image with random rotations to one of 1000 images (from VOC dataset \cite{pascal-voc-2007}) randomly. The use of randomization each iteration and non-barcode (0) inputs are made to ensure the model generalization and reduce the false-positive. We often set $250000$ synthetic samples for training from scratch and increase this to tune models.

On license plate reading, we used AOLP dataset \cite{hsu2012application} as its reliability and challenging conditions. The dataset consists of 2049 images of car/motorbike license plates. The samples are collected at different locations, times, traffic densities, and weather conditions. The dataset consists of 3 categories which are Access Control (AC), Law Enforcement (LE), and Road Patrol (RP). Each category has unique characteristics, and LE, RP conditions are tougher to deal with. Specifically, we use the same train-test split as this state-of-the-art paper \cite{chen2019automatic} used: 581 samples of AC, 657 samples of LE, 511 of RP for training while 300 items (100 each) from AC, LE and RP are used for evaluation. Synthesize samples have similar features as the real set: $3-6$-digit sequences with dash (-) symbol in between, $10$ plate templates with random cut (for obscured $3-4$-digit plate), center-aligned, various colors, and various scale.

\subsection{Experimental Setups}

We chose Dynamsoft, Cognex, Inlite, Zxing and a detect-crop-decode solution (i.e. the step-by-step approach) as competitors in evaluating barcode sets. The first 3 software are trustworthy and used a lot in the industry (thousands of dollars license). While Cognex is evaluated via an online demo website, other official binaries were tested on our computer. No test-time augmentation is used. The step-by-step approach use YOLOv2 \cite{redmon2017yolo9000} with 1 class detection and Pyzbar library, \cite{do2020smart} fastest model for decoding. The training processes were made on Intel Core i7 9700, NVIDIA Titan RTX 24GB from scratch (no pretrain), while evaluations were done on Intel Core i7 4700 and GTX 1060 6GB. We use mostly batch size of $32$ or $16$, and the initial learning rate of $0.01$. We reconfigure these hyper-params and $\lambda_{class}$ to achieve the best results. In decoding the output tensor, confidence threshold and NMS threshold are set at 0.2 and 0.3, respectively. The standard input size is $448 \times 448$, grid size $S=14$, number of boxes $B=2$. For the higher input resolution, we slice it into standard-size patches by a stride, make predictions on patches in parallel, then combine those predictions by patch positions and apply NMS a second time to get final results. Test samples are normalized and padding at the size of $448 \times 896$, while the speed test is conducted at the batch size of $1$ and resolution $448 \times 672$ ($\sim$ $480 \times 640$ VGA) to simulate as practical video input.

The license plate dataset is inputted at $640 \times 640$ for both training and evaluation by only padding operation as we avoid resizing because many numbers on plates already very tiny and distorted characters could change ground-truth such as (B, 8), (4, 9) or (1, I). Other setups and tuning hyper-parameters are nearly the same as the barcode problem.

\begin{table*}[!htbp]
\centering
\caption{Model and Tool results on the Main (2000 single-barcode) test set}
\label{tab:main}
\scalebox{1}{
\begin{tabular}{lllllll} 
\hline
 & backbones & mAP & Recog. Rate & Avg. Time/img & GFLOPs & Params \\ 
\hline
\multirow{8}{*}{Ours} & ResNet50 & \textbf{94.38\%} & \textbf{86.90\%} & 127 ms & 17.23 & 26.98 M \\
 & MobileNet2[:15]\_256 & 91.62\% & \textbf{75.35\%} & \textbf{49 ms}  & \textbf{1.45}  & \textbf{3.2} M \\
 & MobileNet2[:15]\_512 & 85.15\% & 72.10\% & 55 ms & 2.31 & 7.62 M \\
 & MobileNet2[:6]\_256 & 83.46\% & 66.00\% & 67 ms & 7.17 & 2.48 M \\
 & MobileNet2[:6]\_512 & 81.43\% & 61.70\% & 74 ms & 20.81 & 6.83 M \\
 & ResNet50\_no\_Detnet & 84.19\% & 77.80\% & 91 ms & 17.05 & 26.09 M \\
 & MobileNet2\_no\_Detnet & 82.93\% & 65.65\% & 53 ms & 1.57 & 3.84 M \\
 & Darknet19 & 87.17\% & 69.65\% & 87 ms & 17.13 & 23.71 M \\ 
\hline
\multirow{4}{*}{Tools} & Zxing & 26.90\% & 26.90\% & \textbf{53 ms}  & - & - \\
 & Dynamsoft & \textbf{76.55\%} & \textbf{76.30\%} & 375 ms & - & - \\
 & Cognex & 67.15\% & 66.75\% & 115 ms & - & - \\
 & Inlite & 51.35\% & 50.55\% & 218 ms & - & - \\
\hline
\multirow{2}{*}{\begin{tabular}[c]{@{}l@{}}Detect \& crop \\\& decode solution\end{tabular}} & YOLOv2+ Pyzbar & \textbf{97.24\%} & 58.4\% & \textbf{47 ms} & - & - \\
 & YOLOv2+\cite{do2020smart}'s fastest & \textbf{97.24\%} & \textbf{74.8\%} & 75 ms & - & - \\
\hline
\end{tabular}}
\end{table*}

\subsection{Evaluation}
In this work, we adopt: standard Mean Average Precision (mAP with threshold $0.5$) to evaluate detection performance; recognition $rate$ to measure the accuracy of predict each entire sequence correctly, formally 
\[\frac{\# \: of \: correct \: decoded}{\# \: of \: total \: barcodes}; \]
prediction time (millisecond) per one input; the number of floating-point operations (FLOPs), and the number of parameters that model has (i.e., how lightweight the model is).

Table \ref{tab:main} shows us the best models of each backbone and industrial tools. MobileNet2[:X]\_Y means we take only a part of MobileNetV2 from layer $X^{th}$ and stick Det blocks with $Y$ channels inside. Clearly, the ResNet backbone takes the first place but relatively slow with a bigger model size. On the other hand, MobileNet2[:15]\_256 not only gets excellent result even higher all tools but also with a lightning inferencing speed whereas its shallower variants MobileNet2[:6]\_256 and MobileNet2[:6]\_512 took too low abstraction features leading to bad performances. We also witness how useful Det blocks (first 5 models use) since models without them (\_no\_Detnet) could not soar higher. However, when we tried to increase the number of channels inside Det blocks, it does not help, so we have to be careful in setting this number (256 is the default setting used in \cite{li2018detnet}). As we anticipated before, the Darknet19 backbone (YOLOv2 \cite{redmon2017yolo9000}'s backbone, no Det blocks) did fairly well in detection; however, it did not demonstrate adequate performance in recognition because of object smallness in this dataset. Similarly, the detect-crop-decode way did best on detection but poor on decoding by a weak decoder, or decent however slower. Last but not least, we also used MobileNet2[:15]\_256 to do speed evaluation on 2000 EAN-13 test set at a resolution of $448\times672$ ($\sim 480\times640$ VGA), and the result we got is 25.8 ms/image $\sim$ a smooth and real-time 38.76 FPS.

To demonstrate the ability of the model under multiple barcode images and the extensibility of adding a new type of barcode, we also conducted an evaluation on the multi-barcode dataset and on the EAN-8 barcode set as results in Table \ref{tab:multi}. On the multiple set, our best model (ResNet50 with Detnet) get 95.98\% mAP and 81.8\% for accuracy, which is very good and outperformed the commercial counterpart with only 63.6\% even though we set the maximum effort in the setting. The extensibility of this method is also proved by the winning of MobileNet2[:15]\_256 over Zxing (the fastest commercial tool) after training purely on synthetic samples but still achieved 81.1\% on 90 EAN-8 test samples.

\begin{table}[ht]
\centering
\caption{Multi-barcode and EAN-8 evaluation}
\label{tab:multi}
\begin{tabular}{clll} 
\hline
\multicolumn{1}{l}{} & \multicolumn{1}{c}{Model} & mAP & Recog. Rate \\ 
\hline
\multirow{2}{*}{Multi set} & ResNet50 (our best) & \textbf{95.98}\%  & \textbf{63/77} $\sim$ \textbf{81.8}\%  \\
 & Dynamsoft (com. best) & 68.7\% & 49/77 $\sim$ 63.64\%  \\ 
\hline
\multirow{2}{*}{EAN-8 set} & MobileNet2 (our fastest) & \textbf{87.99}\%  & \textbf{73/90} $\sim$ \textbf{81.1}\%  \\
 & Zxing (com. fastest) & 45.56\% & 41/90 $\sim$ 45.56\%  \\
\hline
\end{tabular}
\end{table}

On the work of license plate recognition task, we conducted experiments on 2 backbones which are ResNet50 and MobileNet2[:15] with normal DetNet blocks. As we mentioned about 3 categories of AOLP dataset before, besides the overall evaluation, we separately measure each category with an aim to see how the models react to each condition, where model showed weakness to improve in feature works. As we can see in Table \ref{tab:aolp_resu}, the ResNet-backed model again took the winner position in both mAP (99.35\%) figures and recognition rate (92.21\%) overall. Although the lightweight MobileNetV2-backed model comes second in mAP, it not only did a very good job in term of recognition rate but also the fastest model with only 52.54 ms per image compared with the state-of-the-art paper (SOTA) \cite{chen2019automatic} on this dataset (\emph{note that \cite{chen2019automatic} setup is not the same with ours -- we used GTX 1060 which is only 10\% faster than GTX 970 \cite{chen2019automatic} used}). Last but not least, both of our models do inference very fast although it is still not real-time rate (25-30 fps). As expected, LE and RP are challenging conditions, so our model predicted less correctly in both detection and full-plate classification. We might need to learn more about those conditions and provide better data augmentations that fit them.

\begin{table}[ht]
\centering
\caption{Results on AOLP dataset}
\label{tab:aolp_resu}
\scalebox{1.1}{
\begin{tabular}{llrrr} 
\hline
 &  & \multicolumn{1}{l}{ResNet50} & \multicolumn{1}{l}{MobileNet2} & \multicolumn{1}{l}{\cite{chen2019automatic}} \\ 
\hline
\multirow{2}{*}{AC} & mAP & 100\% & 100\% & - \\
 & recog. rate & 97.00\% & 95.00\% & - \\ 
\hline
\multirow{2}{*}{LE} & mAP & 98.11\% & 98.15\% & - \\
 & recog. rate & 91.67\% & 87.96\% & - \\ 
\hline
\multirow{2}{*}{RP} & mAP & 100.00\% & 99.00\% & - \\
 & recog. rate & 89.00\% & 85.00\% & - \\ 
\hline
\multirow{2}{*}{Overall} & mAP & \textbf{99.35\%}  & 99.02\% & 99.00\% \\
 & recog. rate & \textbf{92.21\%}  & 89.29\% & 78.00\% \\ 
\hline
\multicolumn{2}{l}{GFLOPs} & 35.13 & \textbf{2.93}  & - \\ 
\hline
\multicolumn{2}{l}{Params (M)} & 27.2 & \textbf{3.4}  & - \\ 
\hline
\multicolumn{2}{l}{Time (ms)} & 70.86 & \textbf{52.54}  & 800 - 1000 \\
\hline
\end{tabular}}
\end{table}

\subsection{Ablation Study}

Regarding the influence of different data augmentations, as Table \ref{tab:ablation_0} shown, it is obvious that the shearing effect and rotation are crucial for this model training because, without it, the model dim down clearly in recognition rate (57.35\% and 78.85\% respectively). On the other hand, training without making blurry, adding Gaussian noise or elastic effects just reduces the model performance a little bit. We guessed that because those conditions exist already in the real dataset. 

Back to the subsection of Training in Methodology when we talked about $\lambda_{class}$, we assume that we should put a small value for that hyper-parameter when we start training from scratch. The reason behind is that model should focus more on learning to know where each object is located rather than classifying what each object. So bigger $\lambda_{class}$ would just dominate loss function and hinders the spatial convergence to the correct grid, and consequently, it does not classify digits of sequence correctly either. Therefore, only after a certain degree of convergence appeared (i.e. the model has learned to locate correctly), we should increase the hyper-parameter value to emphasize the role of classification digits correctly. Empirical results at Table \ref{tab:ablation_0} demonstrate our hypothesis as we can see: no matter by ResNet or MobileNetV2, if we set $\lambda_{class}$ too high from the beginning, the model gets confused and could not converge; however, keeping $\lambda_{class}$ low for the whole process also does not produce superior models; and lastly, rise the hyper-parameter value higher ($5$ after $1$) after certain convergence results in better models.

\begin{table}[ht]
\centering
\caption{Data augmentation and $\lambda_{class}$ effect ablation study}
\label{tab:ablation_0}
\scalebox{1.2}{
\begin{tabular}{lrr}
\hline
 & \multicolumn{1}{l}{mAP} & \multicolumn{1}{l}{Recog Rate} \\
\hline
ResNet50\_w1 & 91.87\% & 82.20\% \\
ResNet50\_w5 & 34.28\% & 0.20\% \\
ResNet50\_w1\_5 & \textbf{94.38\%} & \textbf{86.90\%} \\
\hline
MobileNet2[:15]\_256\_w1 & \textbf{96.60\%} & 69.85\% \\
MobileNet2[:15]\_256\_w5 & 46.33\% & 0.00\% \\
MobileNet2[:15]\_256\_w1\_5 & 91.62\% & \textbf{75.35\%} \\
\hline
ResNet50\_w1\_no\_elastic & 92.95\% & 80.95\% \\
ResNet50\_w1\_no\_rotate & \textbf{86.62\%} & \textbf{57.90\%} \\
ResNet50\_w1\_no\_shear & 93.85\% & 78.85\% \\
ResNet50\_w1\_no\_blur & 91.23\% & 81.15\% \\
ResNet50\_w1\_no\_noise & 94.38\% & 81.55\% \\
\hline
\end{tabular}}
\end{table}

\begin{figure}[ht]
\centering
  \includegraphics[width=\linewidth,keepaspectratio]{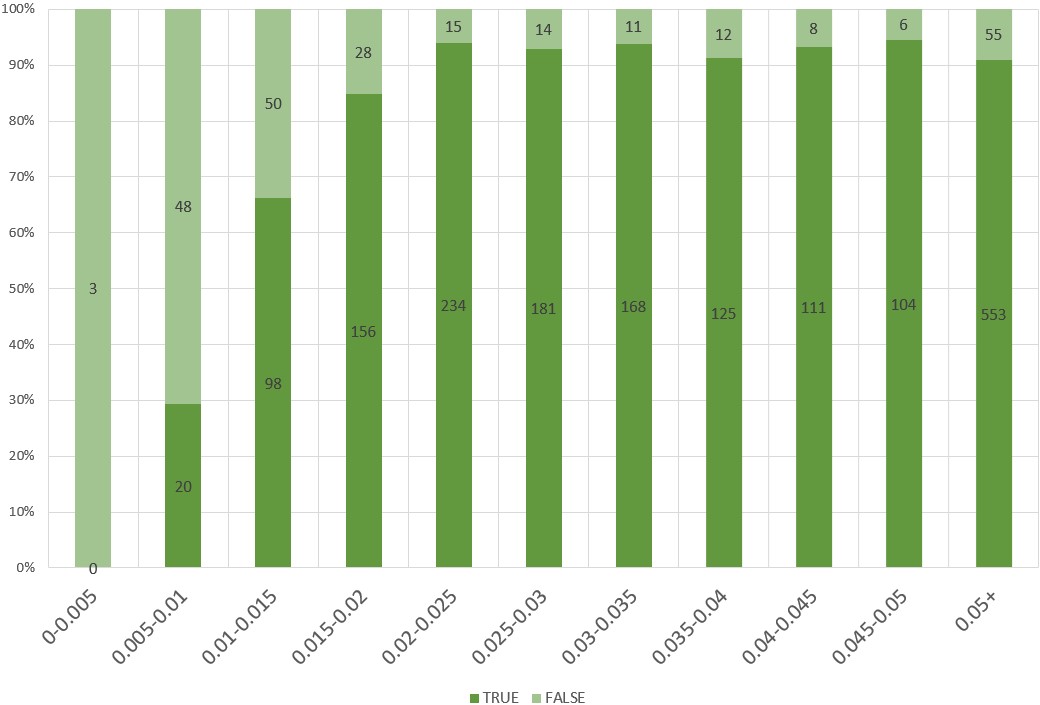}
  \caption{Barcode area on image ratio and decodability}
  \label{fig:ablation_2}
\end{figure}

Eventually, during researching this topic, we always keep in mind that in many cases, even we humans sometimes found and absolutely sure that a certain small object in an image is something we know. However, we cannot see clearly what on. That could also happen to the machine, so we tried to analyze correct decoded and non-decodable (and wrongly predicted included) cases to answer the question: How small of a barcode in an image could make a model difficult to decode correctly ? The answer to this also helps us to notice that if a barcode is too small compared to the whole image, we should be careful with the model outcome, or we had better to make the camera focus closer to the detected location to have a better and trustworthy prediction. As Figure \ref{fig:ablation_2}, it is obvious that if the barcode area is larger than 0.02 of the whole image, our best model (ResNet50) is ready to give over 90\% correctly. (This analysis is done on 2000 main test set).

\section{Conclusion}
In conclusion, in this study, we successfully proposed QuickBrowser to quickly detect and read simple object content by integrating multi-digit classification into DNN one-stage object detection with modified architecture, loss function, data augmentation, and training to make it lightweight, fast, and robust even under most real-life conditions.
We also upgraded an existing dataset to a rigorous benchmark dataset for barcode scanning by collecting more samples, annotating, and double-checking. The dataset has a main EAN-13 single-barcode set having a clear train/validation/test split, an EAN-13 multi-barcode set, and an EAN-8 set to confirm the model extensibility. The dataset is public to promote further research. Lastly, our experimental results on the extended barcode dataset and the AOLP license plate dataset demonstrated the method's efficiency by outperforming industrial software, existing works in terms of detection rate, recognition rate, and inference duration. At the same time, our fastest models accomplish real-time and near-real-time rates under realistic resolutions.

However, our methodology also has a few limitations, such as grid limitation, max length of sequence limitation, and the simple object structure. We are still planning for better improvements by more in-depth analysis of datasets for those drawbacks in future works.

\section{Acknowledgements}
This research was supported by the MSIT (Ministry of Science and ICT), Korea, under the Grand Information Technology Research Center support program (IITP-2021-2020-0-01489) supervised by the IITP (Institute for Information \& communications Technology Planning \& Evaluation). It is also supported by Smart City R\&D project of the Korea Agency for Infrastructure Technology Advancement (KAIA) grant funded by the Ministry of Land, Infrastructure and Transport (MOLIT), Ministry of Science and ICT (MSIT) (Grant 21NSPS-B149400-04).

\bibliographystyle{IEEEtran}
\bibliography{IEEEabrv,IEEEexample}

\begin{thebibliography}{10}
\providecommand{\url}[1]{#1}
\csname url@samestyle\endcsname
\providecommand{\newblock}{\relax}
\providecommand{\bibinfo}[2]{#2}
\providecommand{\BIBentrySTDinterwordspacing}{\spaceskip=0pt\relax}
\providecommand{\BIBentryALTinterwordstretchfactor}{4}
\providecommand{\BIBentryALTinterwordspacing}{\spaceskip=\fontdimen2\font plus
\BIBentryALTinterwordstretchfactor\fontdimen3\font minus
  \fontdimen4\font\relax}
\providecommand{\BIBforeignlanguage}[2]{{%
\expandafter\ifx\csname l@#1\endcsname\relax
\typeout{** WARNING: IEEEtran.bst: No hyphenation pattern has been}%
\typeout{** loaded for the language `#1'. Using the pattern for}%
\typeout{** the default language instead.}%
\else
\language=\csname l@#1\endcsname
\fi
#2}}
\providecommand{\BIBdecl}{\relax}
\BIBdecl

\bibitem{joseph1994bar}
E.~Joseph and T.~Pavlidis, ``Bar code waveform recognition using peak
  locations,'' \emph{IEEE Transactions on Pattern Analysis and Machine
  Intelligence}, vol.~16, no.~6, pp. 630--640, 1994.

\bibitem{muniz1999robust}
R.~Muniz, L.~Junco, and A.~Otero, ``A robust software barcode reader using the
  hough transform,'' in \emph{Proceedings 1999 International Conference on
  Information Intelligence and Systems (Cat. No. PR00446)}.\hskip 1em plus
  0.5em minus 0.4em\relax IEEE, 1999, pp. 313--319.

\bibitem{wachenfeld2008robust}
S.~Wachenfeld, S.~Terlunen, and X.~Jiang, ``Robust recognition of 1-d barcodes
  using camera phones,'' in \emph{2008 19th International Conference on Pattern
  Recognition}.\hskip 1em plus 0.5em minus 0.4em\relax IEEE, 2008, pp. 1--4.

\bibitem{fridborn2017reading}
\BIBentryALTinterwordspacing
F.~Fridborn, ``Reading barcodes with neural networks,'' 2017. [Online].
  Available:
  \url{http://liu.diva-portal.org/smash/record.jsf?pid=diva2:1164104}
\BIBentrySTDinterwordspacing

\bibitem{hansen2017real}
D.~K. Hansen, K.~Nasrollahi, C.~B. Rasmussen, and T.~B. Moeslund, ``Real-time
  barcode detection and classification using deep learning.'' in \emph{IJCCI},
  2017, pp. 321--327.

\bibitem{lin2011real}
D.-T. Lin, M.-C. Lin, and K.-Y. Huang, ``Real-time automatic recognition of
  omnidirectional multiple barcodes and dsp implementation,'' \emph{Machine
  Vision and Applications}, vol.~22, no.~2, pp. 409--419, 2011.

\bibitem{katona2012novel}
M.~Katona and L.~G. Ny{\'u}l, ``A novel method for accurate and efficient
  barcode detection with morphological operations,'' in \emph{2012 Eighth
  International Conference on Signal Image Technology and Internet Based
  Systems}.\hskip 1em plus 0.5em minus 0.4em\relax IEEE, 2012, pp. 307--314.

\bibitem{soros2013blur}
G.~S{\"o}r{\"o}s and C.~Fl{\"o}rkemeier, ``Blur-resistant joint 1d and 2d
  barcode localization for smartphones,'' in \emph{Proceedings of the 12th
  International Conference on Mobile and Ubiquitous Multimedia}, 2013, pp.
  1--8.

\bibitem{creusot2015real}
C.~Creusot and A.~Munawar, ``Real-time barcode detection in the wild,'' in
  \emph{2015 IEEE winter conference on applications of computer vision}.\hskip
  1em plus 0.5em minus 0.4em\relax IEEE, 2015, pp. 239--245.

\bibitem{creusot2016low}
------, ``Low-computation egocentric barcode detector for the blind,'' in
  \emph{2016 IEEE International Conference on Image Processing (ICIP)}.\hskip
  1em plus 0.5em minus 0.4em\relax IEEE, 2016, pp. 2856--2860.

\bibitem{safaei2016real}
A.~Safaei, H.~L. Tang, and S.~Sanei, ``Real-time search-free multiple license
  plate recognition via likelihood estimation of saliency,'' \emph{Computers \&
  Electrical Engineering}, vol.~56, pp. 15--29, 2016.

\bibitem{kumar2016efficient}
T.~Kumar, S.~Gupta, and D.~S. Kushwaha, ``An efficient approach for automatic
  number plate recognition for low resolution images,'' in \emph{Proceedings of
  the Fifth International Conference on Network, Communication and Computing},
  2016, pp. 53--57.

\bibitem{chen2019automatic}
R.-C. Chen \emph{et~al.}, ``Automatic license plate recognition via
  sliding-window darknet-yolo deep learning,'' \emph{Image and Vision
  Computing}, vol.~87, pp. 47--56, 2019.

\bibitem{do2020smart}
T.~Do, Y.~Tolcha, T.~J. Jun, and D.~Kim, ``Smart inference for multidigit
  convolutional neural network based barcode decoding,'' \emph{arXiv preprint
  arXiv:2004.06297}, 2020.

\bibitem{zamberletti2010decoding}
A.~Zamberletti, I.~Gallo, M.~Carullo, and E.~Binaghi, ``Decoding 1-d barcode
  from degraded images using a neural network,'' in \emph{International
  Conference on Computer Vision, Imaging and Computer Graphics}.\hskip 1em plus
  0.5em minus 0.4em\relax Springer, 2010, pp. 45--55.

\bibitem{yang2016automatic}
H.~Yang, L.~Chen, Y.~Chen, Y.~Lee, and Z.~Yin, ``Automatic barcode recognition
  method based on adaptive edge detection and a mapping model,'' \emph{Journal
  of Electronic Imaging}, vol.~25, no.~5, p. 053019, 2016.

\bibitem{goodfellow2013multi}
I.~J. Goodfellow, Y.~Bulatov, J.~Ibarz, S.~Arnoud, and V.~Shet, ``Multi-digit
  number recognition from street view imagery using deep convolutional neural
  networks,'' \emph{arXiv preprint arXiv:1312.6082}, 2013.

\bibitem{anagnostopoulos2008license}
C.-N.~E. Anagnostopoulos, I.~E. Anagnostopoulos, I.~D. Psoroulas, V.~Loumos,
  and E.~Kayafas, ``License plate recognition from still images and video
  sequences: A survey,'' \emph{IEEE Transactions on intelligent transportation
  systems}, vol.~9, no.~3, pp. 377--391, 2008.

\bibitem{sheng2009real}
H.~Sheng, C.~Li, Q.~Wen, and Z.~Xiong, ``Real-time anti-interference location
  of vehicle license plates using high-definition video,'' \emph{IEEE
  Intelligent Transportation Systems Magazine}, vol.~1, no.~4, pp. 17--23,
  2009.

\bibitem{hsu2012application}
G.-S. Hsu, J.-C. Chen, and Y.-Z. Chung, ``Application-oriented license plate
  recognition,'' \emph{IEEE transactions on vehicular technology}, vol.~62,
  no.~2, pp. 552--561, 2012.

\bibitem{deb2008hsi}
K.~Deb and K.-H. Jo, ``Hsi color based vehicle license plate detection,'' in
  \emph{2008 International Conference on Control, Automation and
  Systems}.\hskip 1em plus 0.5em minus 0.4em\relax IEEE, 2008, pp. 687--691.

\bibitem{girshick2014rich}
R.~Girshick, J.~Donahue, T.~Darrell, and J.~Malik, ``Rich feature hierarchies
  for accurate object detection and semantic segmentation,'' in
  \emph{Proceedings of the IEEE conference on computer vision and pattern
  recognition}, 2014, pp. 580--587.

\bibitem{liu2016ssd}
W.~Liu, D.~Anguelov, D.~Erhan, C.~Szegedy, S.~Reed, C.-Y. Fu, and A.~C. Berg,
  ``Ssd: Single shot multibox detector,'' in \emph{European conference on
  computer vision}.\hskip 1em plus 0.5em minus 0.4em\relax Springer, 2016, pp.
  21--37.

\bibitem{redmon2016you}
J.~Redmon, S.~Divvala, R.~Girshick, and A.~Farhadi, ``You only look once:
  Unified, real-time object detection,'' in \emph{Proceedings of the IEEE
  conference on computer vision and pattern recognition}, 2016, pp. 779--788.

\bibitem{liu2020deep}
L.~Liu, W.~Ouyang, X.~Wang, P.~Fieguth, J.~Chen, X.~Liu, and
  M.~Pietik{\"a}inen, ``Deep learning for generic object detection: A survey,''
  \emph{International journal of computer vision}, vol. 128, no.~2, pp.
  261--318, 2020.

\bibitem{redmon2017yolo9000}
J.~Redmon and A.~Farhadi, ``Yolo9000: better, faster, stronger,'' in
  \emph{Proceedings of the IEEE conference on computer vision and pattern
  recognition}, 2017, pp. 7263--7271.

\bibitem{redmon2018yolov3}
------, ``Yolov3: An incremental improvement,'' \emph{arXiv preprint
  arXiv:1804.02767}, 2018.

\bibitem{lin2014microsoft}
T.-Y. Lin, M.~Maire, S.~Belongie, J.~Hays, P.~Perona, D.~Ramanan,
  P.~Doll{\'a}r, and C.~L. Zitnick, ``Microsoft coco: Common objects in
  context,'' in \emph{European conference on computer vision}.\hskip 1em plus
  0.5em minus 0.4em\relax Springer, 2014, pp. 740--755.

\bibitem{he2016deep}
K.~He, X.~Zhang, S.~Ren, and J.~Sun, ``Deep residual learning for image
  recognition,'' in \emph{Proceedings of the IEEE conference on computer vision
  and pattern recognition}, 2016, pp. 770--778.

\bibitem{sandler2018mobilenetv2}
M.~Sandler, A.~Howard, M.~Zhu, A.~Zhmoginov, and L.-C. Chen, ``Mobilenetv2:
  Inverted residuals and linear bottlenecks,'' in \emph{Proceedings of the IEEE
  conference on computer vision and pattern recognition}, 2018, pp. 4510--4520.

\bibitem{yu2017dilated}
F.~Yu, V.~Koltun, and T.~Funkhouser, ``Dilated residual networks,'' in
  \emph{Proceedings of the IEEE conference on computer vision and pattern
  recognition}, 2017, pp. 472--480.

\bibitem{li2018detnet}
Z.~Li, C.~Peng, G.~Yu, X.~Zhang, Y.~Deng, and J.~Sun, ``Detnet: Design backbone
  for object detection,'' in \emph{Proceedings of the European conference on
  computer vision (ECCV)}, 2018, pp. 334--350.

\bibitem{simard2003best}
P.~Y. Simard, D.~Steinkraus, J.~C. Platt \emph{et~al.}, ``Best practices for
  convolutional neural networks applied to visual document analysis.'' in
  \emph{Icdar}, vol.~3, 2003.

\bibitem{lazzari2018foodrepo}
G.~Lazzari, Y.~Jaquet, D.~J. Kebaili, L.~Symul, and M.~Salath{\'e}, ``Foodrepo:
  An open food repository of barcoded food products,'' \emph{Frontiers in
  nutrition}, vol.~5, p.~57, 2018.

\bibitem{pascal-voc-2007}
M.~Everingham, L.~Van~Gool, C.~K.~I. Williams, J.~Winn, and A.~Zisserman, ``The
  {PASCAL} {V}isual {O}bject {C}lasses {C}hallenge 2007 {(VOC2007)}
  {R}esults,''
  http://www.pascal-network.org/challenges/VOC/voc2007/workshop/index.html.

\end{thebibliography}

\end{document}